\documentclass[10pt,twocolumn,letterpaper]{article}

\usepackage{iccv}
\usepackage{times}
\usepackage{epsfig}
\usepackage{graphicx}
\usepackage{amsmath}
\usepackage{amssymb}

\usepackage[breaklinks=true,bookmarks=false]{hyperref}

\iccvfinalcopy 

\def\iccvPaperID{****} 

\ificcvfinal\fi

\begin{document}

\title{Dynamic Attention Networks for Task Oriented Grounding}
\author{ {Soumik Dasgupta} \quad {Badri N. Patro} \quad  {Vinay P. Namboodiri} \\
Indian Institute of Technology, Kanpur \\
{\tt\small \{soumikdasguptasl@gmail.com\}, \{ badri,vinaypn \}@iitk.ac.in}
}

\maketitle
\ificcvfinal\thispagestyle{empty}\fi


\begin{abstract}
In order to successfully perform tasks specified by natural language instructions, an artificial agent operating in a visual world needs to map words, concepts, and actions from the instruction to visual elements in its environment. This association is termed as Task-Oriented Grounding. In this work, we propose a novel Dynamic Attention Network architecture for the efficient multi-modal fusion of text and visual representations which can generate a robust definition of state for the policy learner. Our model assumes no prior knowledge from visual and textual domains and is an end to end trainable. For a 3D visual world where the observation changes continuously, the attention on the visual elements tends to be highly co-related from one-time step to the next. We term this as "Dynamic Attention". In this work, we show that Dynamic Attention helps in achieving grounding and also aids in the policy learning objective. Since most practical robotic applications take place in the real world where the observation space is continuous, our framework can be used as a generalized multi-modal fusion unit for robotic control through natural language. 
We show the effectiveness of using 1D convolution over Gated Attention Hadamard product on the rate of convergence of the network. We demonstrate that the cell-state of a Long Short Term Memory (LSTM) is a natural choice for modeling Dynamic Attention and show through visualization that the generated attention is very close to how humans tend to focus on the environment.
\end{abstract}

\section{Introduction}
\begin{figure}[htb]
 \small
 \centering
 \begin{tabular}[b]{ c }
 (a) Go to the short green pillar  Network Objective \\ 
 \includegraphics[width=0.44\textwidth]{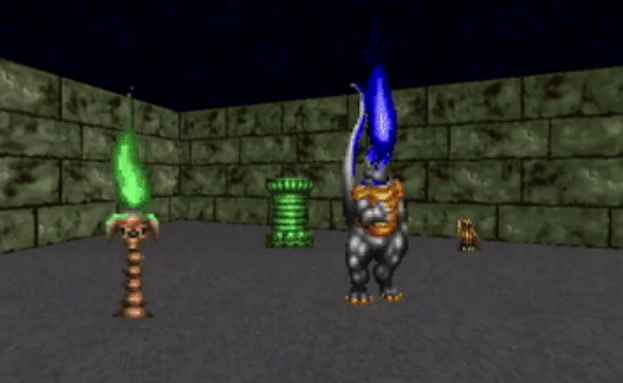}\\ 
  (b) Dynamic Attention\\\\
 \includegraphics[width=0.44\textwidth]{overall_pipeline.png}\\
   \end{tabular}
  \caption{ This figure indicate the overview of the problem.}
  \label{fig:intro}
  \vspace{-1em}
\end{figure}

To have AI systems navigate and carry out instructions in a visual world, an agent needs to extract semantically meaningful representation of natural language by mapping it to visual elements in the environment. We simulate the
 problem of Task Oriented Grounding by training an agent to take natural language instructions and learn to navigate a virtual environment introduced by \cite{chaplot2017gated}. Consider a scenario as depicted in Fig. \ref{fig:intro}, which shows the egocentric view that the agent sees at some time step.

 The agent receives a natural language instruction at the beginning of every episode and pixel level visual information at every time step, based on which it needs to carry out a navigational task specified by the instruction. To carry out the task with high accuracy, the agent has to draw semantic correspondences between the visual and textual modalities in order to learn a policy. This problem has several challenges: 1) The agent has to have the ability to recognize the objects indicated by the instructions, 2) It needs to have some notion of memory of the previous observations in order to explore the environment since the object concerned may not be in the field of view, 3) It has to ground each concept of the instruction in the environment and reason about the semantics, eg. instructions having superlative degree - `Go to the tallest torch'. and 4) It has to learn a policy so that it can successfully navigate to the correct object while avoiding the incorrect ones. The main contribution of the work is a state processing module that uses a Dynamic Attention Architecture for multi-modal fusion to generate an informative and robust definition of state for the policy learning module to see. To demonstrate the significance of our Dynamic Attention Network we use the Gated Attention model of \cite{chaplot2017gated} as the baseline. For fair comparison we adopt the same overall architecture and the environment settings from the baseline.

The main contribution of this paper are follows:
\begin{itemize}
    \item We propose a novel Dynamic Attention Network for generating attention to improve response in Task Oriented Language Grounding.
    \item We propose 1D convolution as a method for multi-modal fusion over the Hadamard product to achieve faster convergence of the network.
    \item We demonstrate experimental results to show the effects of Dynamic Attention on the accuracy and convergence rate with various similar architectures that differ subtly but produces significant changes when it comes to the overall performance. 
  \item We show visualizations of the attention masks generated by Dynamic Attention Network(DAN) to demonstrate its robustness. We compare Zero-Shot (ZS) and Multi-Task (MT) generalization accuracy with the baseline for three modes of difficulty of the task.
\end{itemize}

\begin{figure*}[htb]
\centering
\includegraphics[width=1\textwidth]{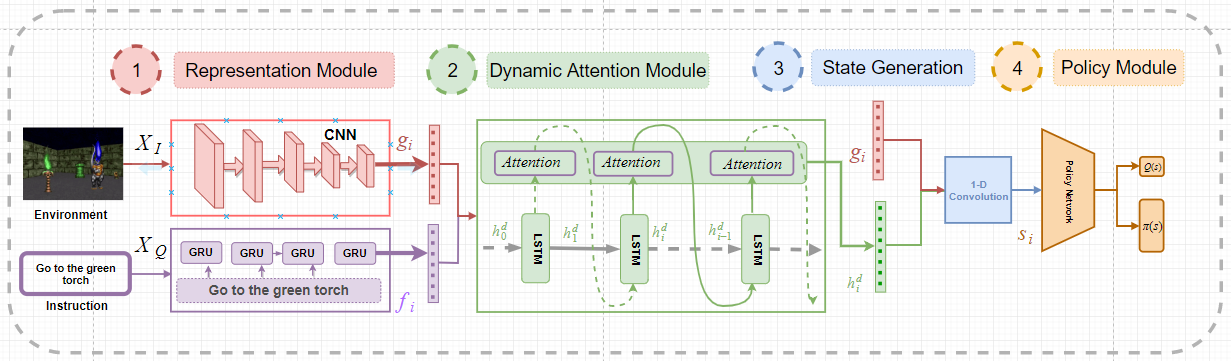}
\caption{\textit{Overall Architecture}}
\label{fig:ov}
\end{figure*}

\section{Related Work}
The task of grounding is well studied in the various field computer vision and a natural language processing. Staring from image description ~\cite{Barnard_JMLR2003,Farhadi_ECCV2010,Kulkarni_CVPR2011} where natural language concepts are grounded on image. ~\cite{Socher_TACL2014,Vinyals_CVPR2015,Karpathy_CVPR2015,Xu_ICML2015,Fang_CVPR2015,Chen_CVPR2015,Johnson_CVPR2016,Yan_ECCV2016} have generated descriptive sentences from images with the help of Deep Networks. To generate engaging question about image is known as Visual Question Generation (VQG) \cite{Mostafazadeh_ACL2016,jain_CVPR2017,Patro_EMNLP2018MDN}. To generate similar type grounding question given question is known as paraphrase question generation\cite{Patro_COLING2018learning}.  Also, a variety of methods have been proposed  by~\cite{Malinowski_NIPS2014,Lin_ECCV2014,VQA,Ren_NIPS2015,Ma_AAAI2016,Noh_CVPR2016} for grounding natural language question on image for solving visual question answering (VQA) task. For grounding natural language question on image for solving VQA task includes attention-based methods~\cite{Zhu_CVPR2016,Fukui_arXiv2016,Gao_NIPS2015,Xu_ECCV2016,Lu_NIPS2016,Shih_CVPR2016,Patro_CVPR2018dvqa,Patro_ICCV2019}. Also, There have been many works for solving Visual Dialog grounding by asking set of question answering ~\cite{Das_EMNLP2016,Das_CVPR2017,de2017guesswhat,strub2017end,Patro2019ProbabilisticFF}. 

Grounding natural language instructions have been studied in video, such as \cite{Chao}, \cite{Lemaignan2012} look at grounding concepts through human-robot interaction. \cite{Guadarrama}, \cite{Bollini2013} and \cite{Beetz} aimed to ground navigational instruction and the focus was to ground verbs like follow, go, move, pick up etc. \cite{chen:aaai11} learn a navigational policy in a 2D maze like environment by using a semantic parser. \cite{Artzi-2013} and \cite{Misra} ground natural language instruction by mapping instructions to action sequences. \cite{Mei} map navigational instructions to action sequences by representing the state using bag of word features. \cite{DBLP:YuZX17} trained a model to learn to navigate a 2D maze environment. \cite{OhSLK17} study zero-shot generalization in a 3D environment. A similar line of work was done by \cite{Misra} who solve for joint reasoning of linguistic and visual inputs for a task of moving blocks in a 2D environment. They use raw image from the 2D grid, processed by a Convolutional Neural Network (CNN) \cite{LeCunn} and instruction representation obtained through an LSTM \cite{Hochreiter} which are then combined through concatenation.

\cite{chaplot2017gated} propose a Gated-Attention architecture for Task Oriented Language Grounding and evaluate their approach on an environment built
 VizDoom \cite{KempkaWRTJ16}. We use their work as the baseline for comparison. They generate an attention vector as a function of the instruction embedding and fuse with the image representation through a Hadamard product. The problem with this approach is that the attention vector remains static throughout an episode since it is conditioned on the instruction alone. Also it does not leverage the continuity of the observation space of the 3D Doom scenario. In this paper we propose a novel method for multi-modal fusion in the form of Dynamic Attention and show its effectiveness over the the existing benchmarks. Since most of the robotic tasks are in the real world where the observation space is continuous, our model can be a generalized framework in any robotics application.

\section{Environment}

We conduct our experiments on an environment introduced by \cite{chaplot2017gated}. The environment is built on top of the VizDoom API \cite{KempkaWRTJ16}, based on Doom, a classic first person shooter game. The game generates the first person view of the agent at every time step and the agent can interact with the environment by choosing from one of the actions: Turn Right, Turn Left, Move Forward. Each episode starts in a confined room where the agent and various doom objects are spawned at random locations and an instruction of the form "Go to the tall green pillar" is chosen at random from a corpus. The objects have various visual attributes such as color, shape and size. Associated with every instruction there is a set of correct objects.  Each time an instruction is selected, the environment generates a random combination of a correct and 4 incorrect objects and they are spawned at various locations on the map depending on the difficulty. The levels are Easy: Agent is spawned at a fixed location. The objects are spawned in a straight line in the field of view of the agent, Medium: The objects are spawned at random locations even though they are still in the field of view of the agent, Hard: The objects and agent are spawned randomly and the agent can have any initial orientation. The agent might need to explore the environment to see all the objects. The objective of the agent is to navigate to the correct object while avoiding the incorrect ones.

\section{Approach}
\begin{figure*}[htb!]
\centering
\includegraphics[width=1\textwidth]{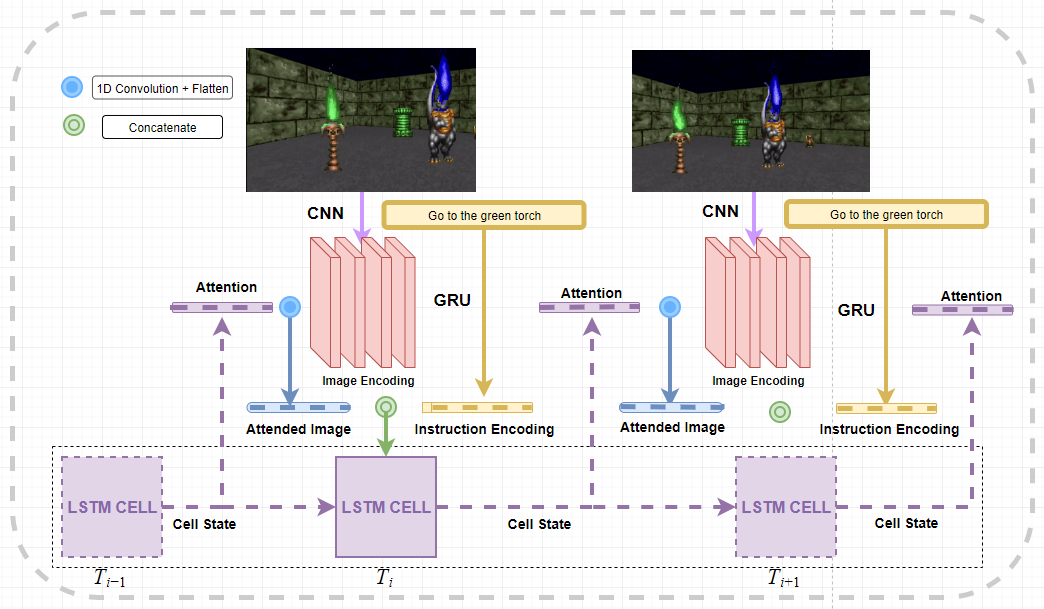}
\caption{\textit{Dynamic Attention Module} The figure shows the Dynamic Attention Module unrolled in time.}
\label{fig:5}
\end{figure*}
Our method consists of four modules as illustrated in Fig~\ref{fig:ov}:

\begin{enumerate}
	\item We obtain embedding for the inputs: current frame image and the instruction using CNN  and Gated Recurrent Unit (GRU)networks. We term this as Representation module.
	\item  We obtain attention vector conditioned on the current frame image and the instruction embedding through the Dynamic Attention Module.
	\item We apply the generated attention vector on the current frame through 1D convolution to generate a state vector.
	\item Finally our policy module will take as input the state and give as output a probability distribution over the action space and a scalar value.
\end{enumerate}

 \subsection{Image Processing Module:} It consists of a 3 layered Convolutional Neural Network \cite{LeCunn} to generate a feature representation of the Image. Let $x_I \in \mathbb{R}^{dxHxW}$ be the feature representation of the Image, $d$ denotes the number of feature maps of the CNN output, $H x W$ is the size of each feature map.

\subsection{Instruction Processing Module:} It consists of a Gated Recurrent Unit (GRU). It encodes the feature representation of the Image as a vector, $x_L \in \mathbb{R}^l$, where l is the dimension of the language encoding.

\subsection{Dynamic Attention Module}
We introduce a Dynamic Attention Network to generate an attention vector over the current image frame. It takes as input the current frame image "attended" with the attention vector of the previous time step concatenated with the instruction encoding and produces the attention vector for the next time step as the output. We model the Dynamic Attention as the Cell-State of an LSTM.

The attention $C_t$ can be formulated as: 
\begin{equation}
    \begin{split}
        & f_t=\sigma(W_f.[h_{t-1}] + b_f)\\
        &i_{t}=\sigma(W_i.[h_{t-1}, x_t] + b_i)\\
        & C'_t = \tanh(W_C.[h_{t-1}, x_t] + b_C) \\
        & C_{t} = f_t*C_{t-1} + i_t*C'_t
    \end{split}
\end{equation}
Where $W_f$, $b_f$, $W_i$, $b_i$, $W_C$, $b_C$ are the weight and bias parameters of the LSTM gates.
Fig. \ref{fig:5} shows our Dynamic Attention Module unrolled over time. At every time-step, the attended image (applying the attention vector generation at the previous time-step) and the instruction representation is concatenated and fed to an LSTM cell as an input. The updated cell-state is the attention vector for the next time step.

\subsection{State Generation Unit}

\begin{figure}[htb]
\centering
\includegraphics[width=0.5\textwidth]{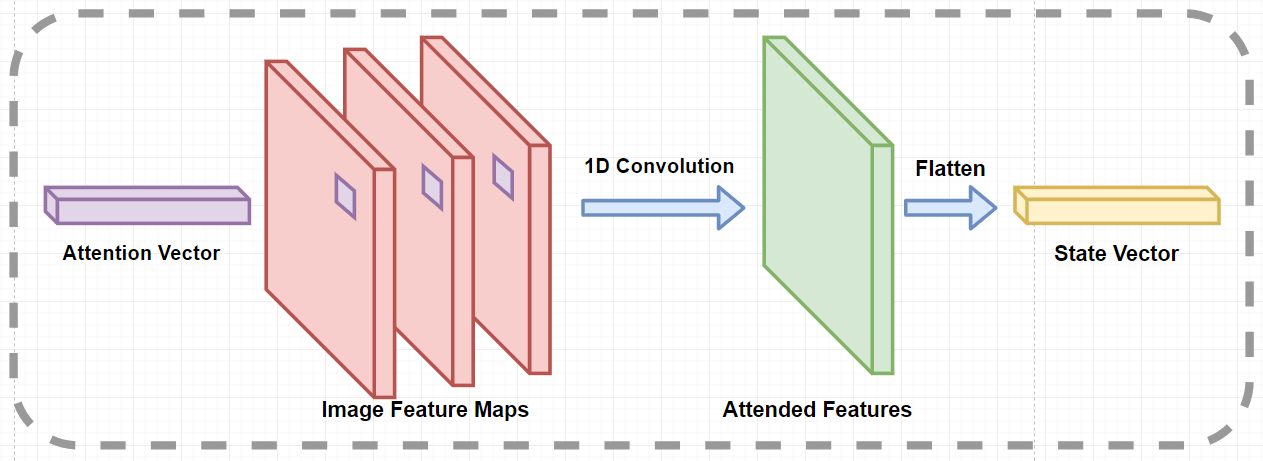}
\caption{\textit{State Generation} The State Representation is generated by applying a 1D convolution of the attention vector on the image feature activation maps}
\label{fig:att_app}
\end{figure}

We generate a state for the policy learner by performing 1D convolution of the attention vector over the feature maps of the current frame as shown in Fig. \ref{fig:att_app}. This is in contrast with \cite{chaplot2017gated} who use Hadamard product for applying the attention vector over the feature maps. We achieve faster convergence rate due to reduced number of parameters of the network.

\begin{figure}[htb]
\label{fig:policy_module}
\centering
\includegraphics[width=0.48\textwidth]{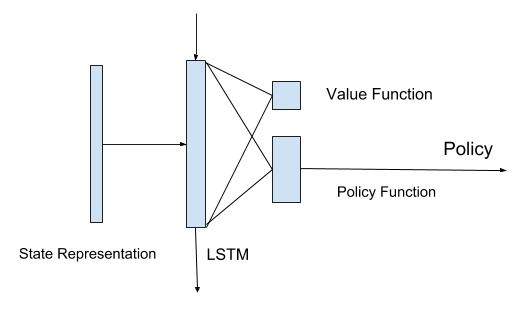}
\caption{\textit{Policy Learning Module} The State Representation is given as an input to a standard A3C learning module that learns a mapping from state to actions}
\label{fig:relatt}
\end{figure}
\subsection{Policy Learning Module}
 The policy learning module is an Asynchronous Advantage Actor Critic (A3C) network \cite{pmlr-v48-mniha16}. The A3C module produces a probability distribution over the action space and a scalar value.The action for the current time step is sampled from the distribution and the action is taken to to get the reward. The policy and value loss is then back-propagated through the network to update the parameters. 
\subsection{Cost Function}
The Actor-Critic algorithms follows an approximate policy gradient:
\begin{equation}
\bigtriangledown_\theta J(\theta) \approx \mathbb{E}_{\pi_\theta}[\bigtriangledown_\theta log \pi_\theta(s,a) Q_w(s, a)]
\end{equation}
Where $Q_w(s,a)$ is the estimate of the value function for taking action $a$ in state $s$. $\pi$ is the policy and $J(\theta)$ is the expected total reward and $\theta$ is the parameters of the policy network.
\begin{table*}[htb]
  \centering
\begin{tabular}[h]{| l |c|c|c|c| c |  c|}
 \hline
 \multicolumn{1}{|c|}{Model}& 
 \multicolumn{3}{|c|}{\textbf{Zero-Shot}} & \multicolumn{3}{|c|}{\textbf{Multi-Task}}
 \\
  \hline
 Mode & Easy & Med & Hard &  Easy & Med & Hard\\
 \hline
 Current Frame attention & 0.964 & 0.912 & 0.817 & 0.969 & 0.926 & 0.791\\
 \hline
 Dynamic Attention LSTM output & 0.981 & 0.952 & 0.852 & 0.966 & 0.959 & 0.842\\
 \hline
 Dynamic Attention LSTM Cell-State & 0.987 & 0.970 & 0.880 & 0.997 & 0.980 & 0.881 \\
 \hline
\end{tabular}
  \caption{\label{tab:ablation} \textit{Ablation Analysis} Comparison of the ZS and MT performances of various models similar to DAN cell-state.}
\end{table*}

\section{Experiment}
We evaluated the proposed method DAN through several experiments and performing quantitative and qualitative analysis. Quantitative analysis includes ablation analysis with similar variants of the model that we tried and analyze the performance of each (Section \ref{AblationQuant}). Comparison of our proposed method with various state of the art models is provided in section \ref{SOTA}. Section \ref{attn_viz} shows qualitative analysis through visualization of the attention maps and study of their properties and failure cases.

\subsection{Environment Setup}

Experiments are performed on all three difficulty modes. During training, the objects are spawned from a training set of 55 instructions and 15 instructions pertaining to unseen attribute-object combinations are held out for a test set for zero-shot evaluation. At each time step the agent is presented with a state definition generated by our state processing module based on which the agent will take one of the three actions. The episode ends if one of the three events occur: The agent reaches an object, the number of time-steps reaches a maximum episode length of T = 30. At the end of each episode the agent receives a reward of: 1 for reaching the correct object, -0.2 for reaching an incorrect object, 0 if the episode times out. Evaluation metric is the accuracy which is the fraction of time the agent reaches the correct object.

The agent is tested on two scenarios suggested by \cite{chaplot2017gated}.
\textbf{Multitask Generalization}: The agent is evaluated on unseen maps having unseen combination of objects at random locations with instructions from the train set. \textbf{Zero-shot Generalization}: The agent is evaluated on unseen test instructions.

\subsection{Implementation Setup}
At every time step, the agent receives the screen buffer image of the environment as a first person view. The image features are extracted using a 3 layer CNN. The image feature has 64 channels each having dimensions 8 x 17. The instruction representation is generated by a GRU of size 256. The attention vector is obtained from the Dynamic Attention Module LSTM and applied to the image by means of a 1D convolution resulting in the attended image representation of size 1 x 8 x 17. The attended image is then flattened in to a vector of size 8 x 17 which gives the state representation that is given as the input to the A3C module as the input. The attention is updated for the next time step by giving as input to the Dynamic Attention LSTM, the concatenated vector of the current state representation and the instruction representation.
The A3C module produces a probability distribution over the action space and a value. Action for this time step is sampled from the distribution and the action taken to get the reward for the time step. The policy and value loss is then back-propagated through the network to update the parameters.

\subsection{Ablation analysis}\label{AblationQuant}
  \begin{figure}[htb] 
   \centering
    \includegraphics[width=0.5\textwidth]{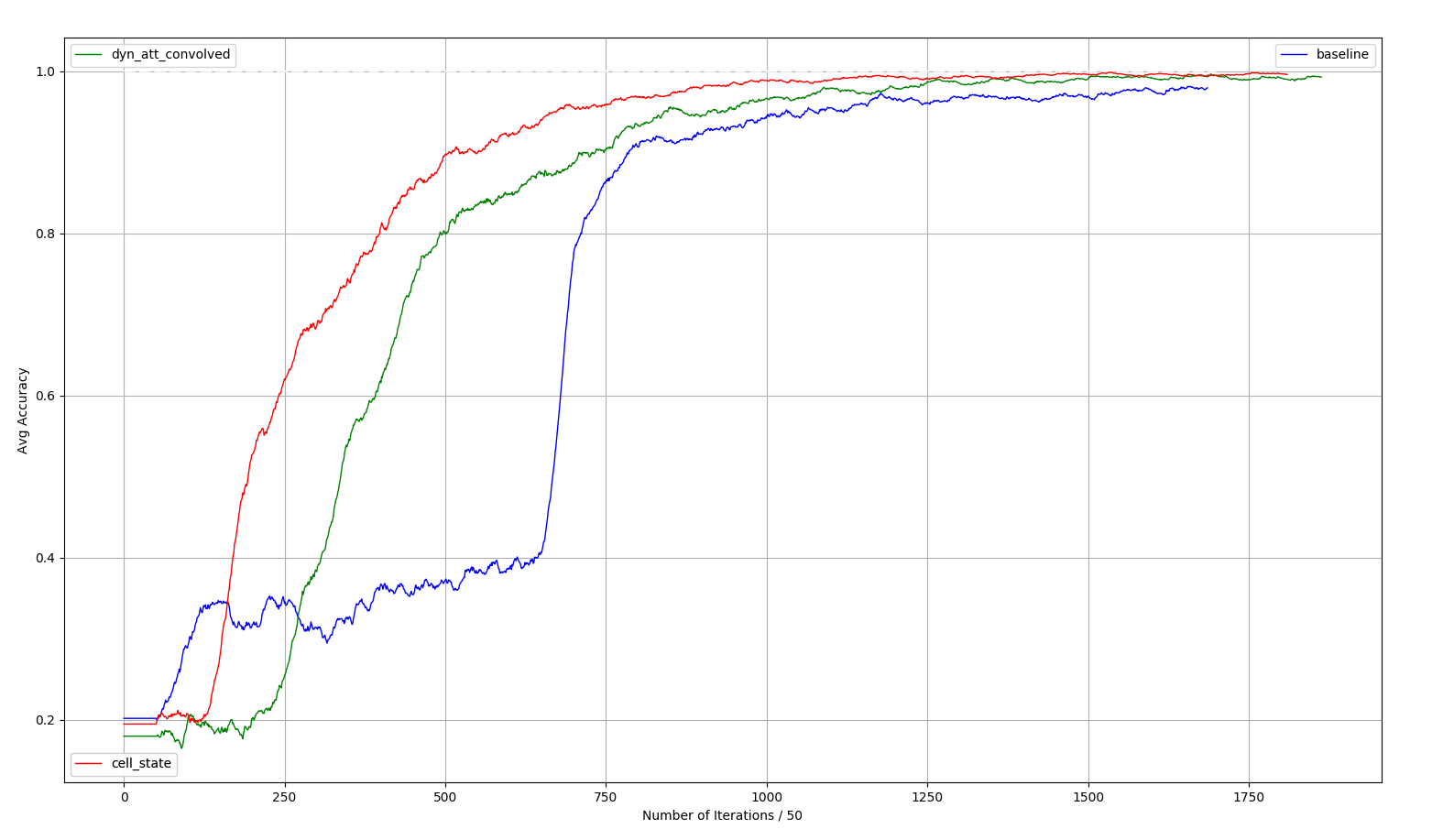}
   \caption{\textit{Dynamic Attention LSTM Cell-State} The figure compares the Dynamic Attention LSTM-output and the Dynamic Attention Cell-State with the baseline}
    \label{fig:6}
\end{figure}
\begin{figure}[]
 \small
 \centering
 \begin{tabular}[b]{ c }
 (a) \textit{Current Frame Attention.} \\ 
 \includegraphics[width=0.5\textwidth]{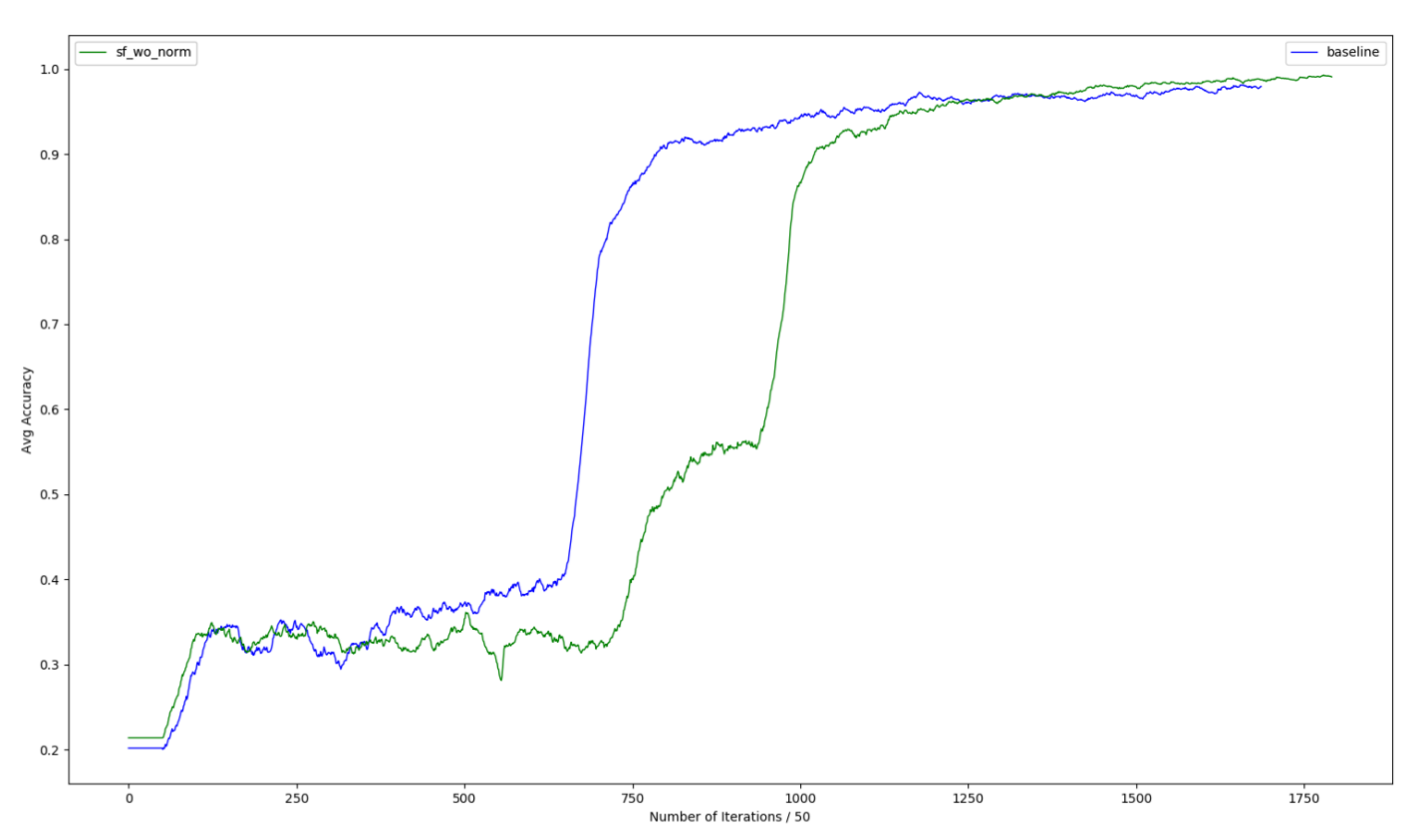}\\
  (b)\textit{Dynamic Attention LSTM-output.} \\
 \includegraphics[width=0.5\textwidth]{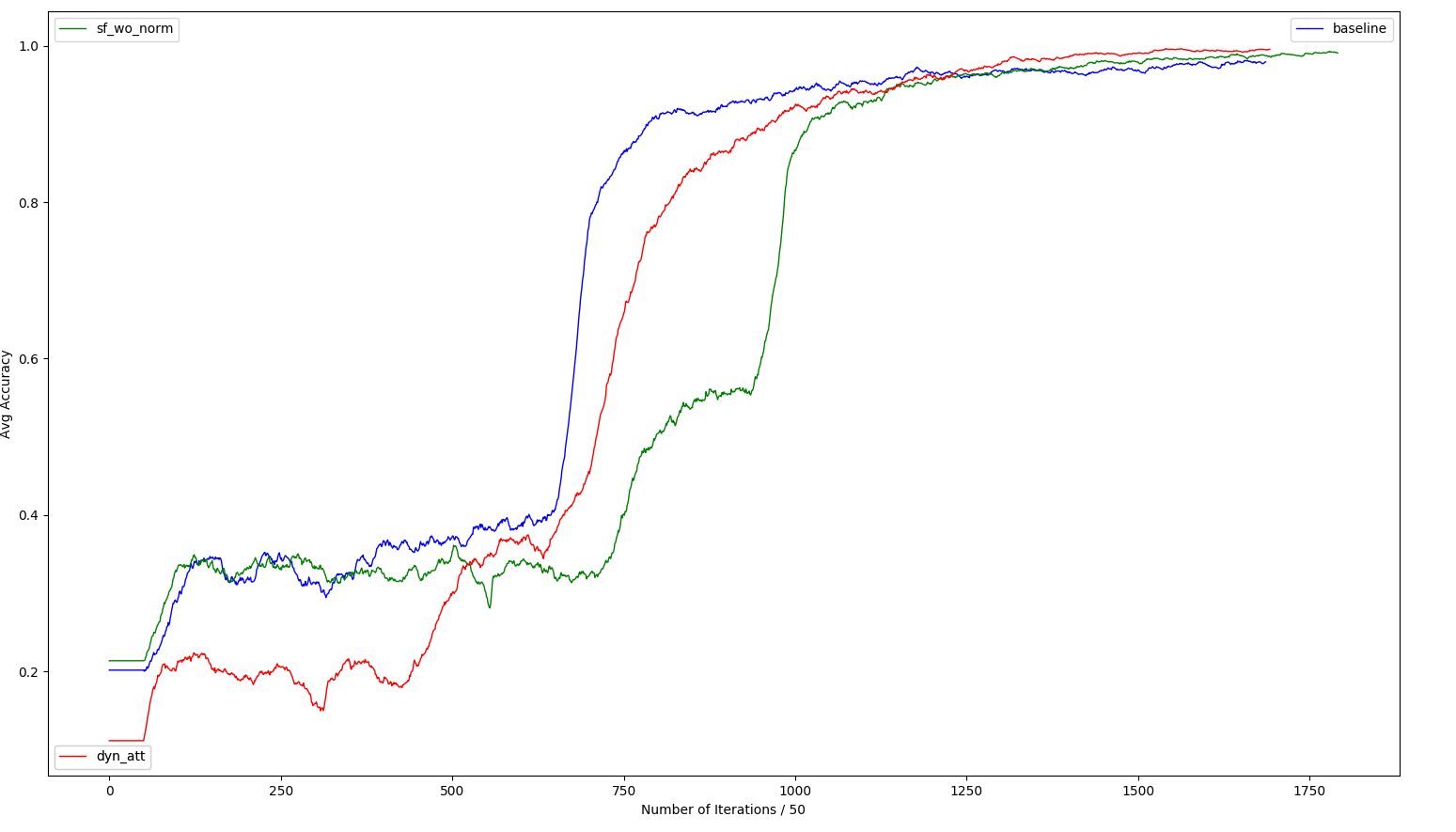}\\
   (c)\textit{1D convolution for faster convergence} \\
 \includegraphics[width=0.5\textwidth]{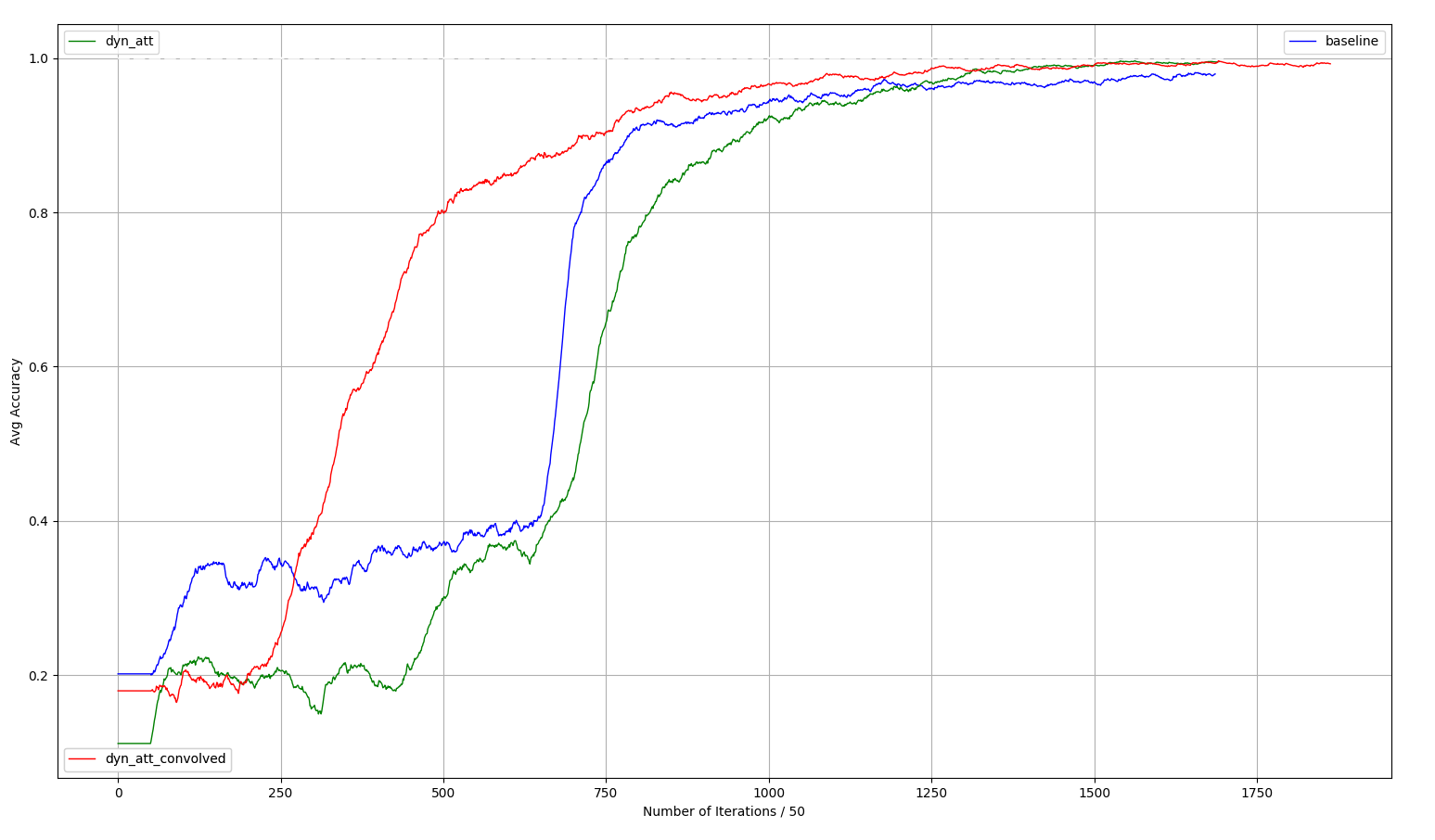}\\
   \end{tabular}
  \caption{(a) The green curve shows training accuracy plot of Current Frame attention and blue is the baseline. (b)The red curve shows the training accuracy plot of the Dynamic Attention - Output model. (c)The red curve shows the faster convergence due to the use of 1D convolution over Gated Attention method. (Blue is the baseline and green is LSTM-output, both using Gated Attention). Accuracy plots of various similar models to Dynamic Attention Network (cell-state) and their comparison }
  \label{fig:f}
  \vspace{-1em}
\end{figure}

For the experiments, we trained each model 3 times from scratch and plotted the mean of their accuracy after each epoch to get the training curve. In the Gated Attention architecture of \cite{chaplot2017gated}, the attention vector is a function of the instruction representation alone. The first hypothesis is that along with the instruction, the current frame image information is also a necessary context for generating the attention. As a proof of concept we use the Gated Attention Network as that of \cite{chaplot2017gated} but make the attention vector a function of concatenation of the instruction encoding and the image convolution features. From Fig. \ref{fig:f}-a, our model shows increased steady state accuracy but slower convergence.

\begin{table}[htb]
  \centering
  \begin{tabular}{|l|c|c|c|}
    \hline
    \textbf{Models}&  \textbf{Easy}& \textbf{Medium} & \textbf{Hard} \\

\hline
Con (\cite{chaplot2017gated}) & 0.928 & 0.680 & 0.280\\ 
\hline
GA (\cite{chaplot2017gated}) & 0.960 & 0.889 & 0.809\\ 
\hline
DAN (our) & \textbf{0.987} & \textbf{0.970} & \textbf{0.880} \\ 
\hline
  \end{tabular}
  \caption{\label{tab_zero_shot} \textit{Comparison with Baseline}Zero Shot Generalization of our method (DAN) with  Gated attention (GA) and concatenation method}
\end{table}

\begin{table}[]
  \centering
  \begin{tabular}{|l|c|c|c|}
    \hline
    \textbf{Models}&  \textbf{Easy}& \textbf{Medium} & \textbf{Hard} \\
    \hline
    Concatenation(\cite{chaplot2017gated})& 0.950 & 0.880 & 0.282\\ 
    \hline
    Gated Attention(\cite{chaplot2017gated}) & 0.958 & 0.964 & 0.825\\
    \hline
    DAN (our) & \textbf{0.997} & \textbf{0.98} & \textbf{0.881} \\ 
    \hline
  \end{tabular}
  \caption{\label{tab_multi_task}This table provides comparison result for  Multi-Task Generalization}
\end{table}
The next hypothesis is that the attention vector at a given time is not independent of those at the previous time steps. The attention vectors of successive time steps should be co-related and need not be computed from scratch every time. To verify, we model the attention as an LSTM. At each time step, we take the attention vector of the previous time step and apply it to the current frame image features. This attended image features concatenated with the instruction encoding is fed to an LSTM cell to generate the attention vector for the next time step. We still use the baseline's Hadamard product for applying the attention on the image features. Fig \ref{fig:f}-b shows the comparison of accuracy plot with current frame and the baseline. We observe a greater steady state accuracy and a faster convergence than the current frame attention.

To tackle the slower convergence rate, we replace the Gated Attention approach by 1D convolution. The baseline uses a Hadamard product followed by downsizing through an FC layer to generate a state definition for the policy learner. The use of 1D convolution eliminates the need for an FC layer to downsize the state representation and reduces the number of trainable parameters. Fig. \ref{fig:f}-c shows the faster convergence rate of 1D convolution over Hadamard product.

 \begin{figure*}[!htb]
 \small
 \centering
 \begin{tabular}[b]{ c }
 (a) Go to the keycard \\ 
 \includegraphics[width=\textwidth]{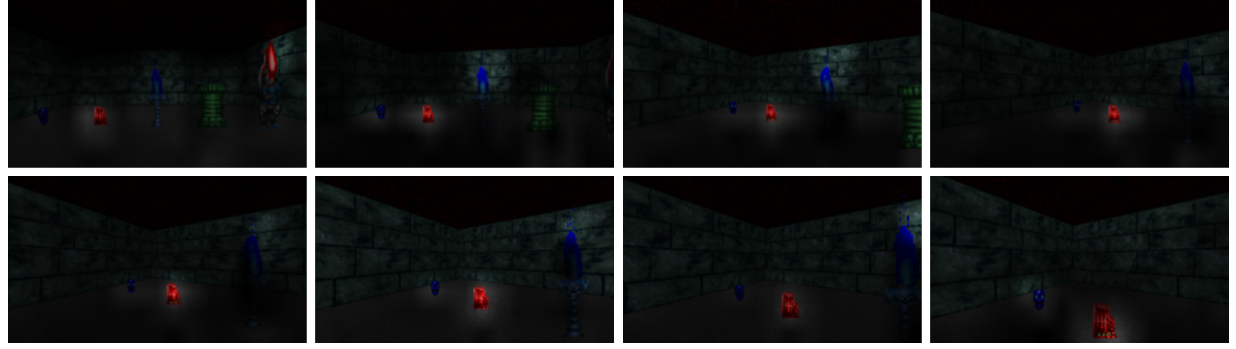}\\
  (b) Go to the tall green pillar\\
 \includegraphics[width=\textwidth]{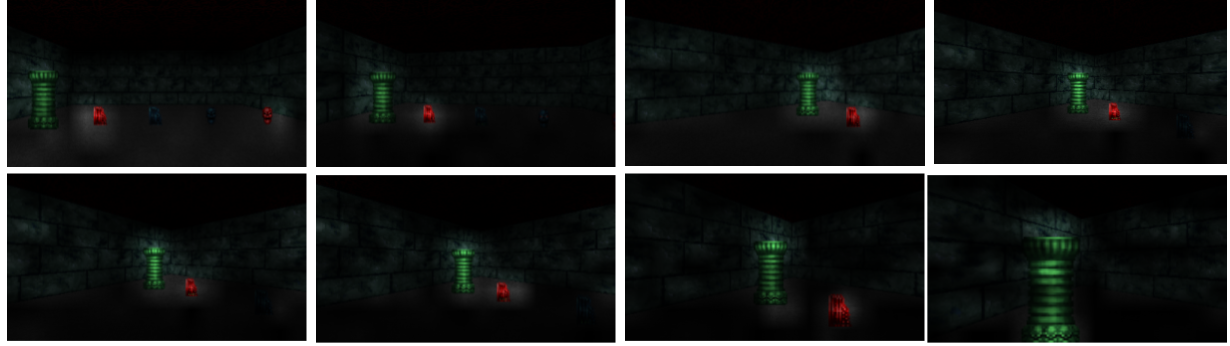}\\
   (c) Go to the red short object\\
 \includegraphics[width=\textwidth]{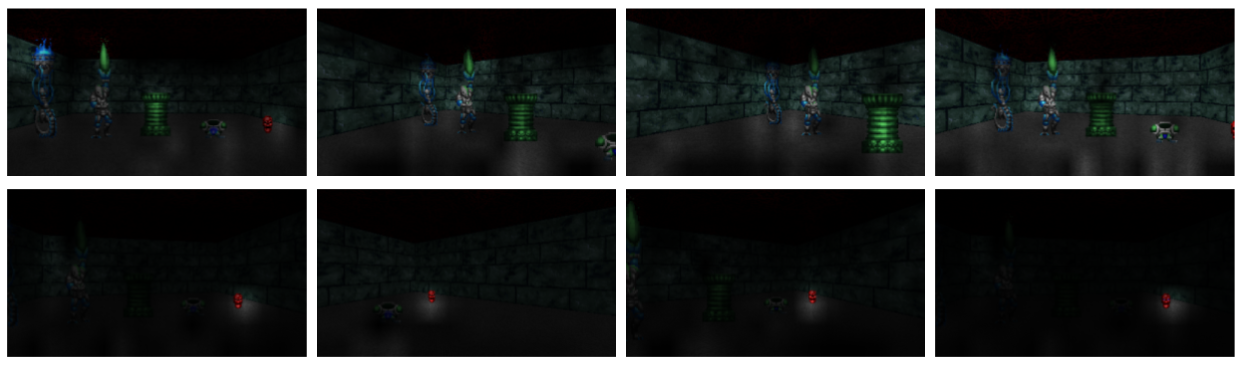}\\
   \end{tabular}
  \caption{ This figure indicates visualization of Attention at regular time steps for three instructions. In figure-(a), the first frame is the reference frames and the last frame indicates the attended frame for a particular instruction.  Similarly figure-(b)  the last frame only green pillar is highlighted and other objects are sub-pressed. In figure-(c), the last frame able to localised the red object among all the object, which follow the instruction carefully.}
  \label{fig:attentions_viz}
  \vspace{-1em}
\end{figure*}

Our next hypothesis is that in continuous observation spaces, the attention tends not to change abruptly. Most of the information is retained in the attention from one time step to the next. Gradually some information is added and removed from the attention vector, as new objects are introduced in the field of view of the agent while some are removed. This behaviour is inherent in the cell-state of an LSTM. Hence to incorporate this inductive bias, we model the attention vector as the cell-state of an LSTM.

Fig \ref{fig:6} shows the accuracy plot of Dynamic Attention LSTM Cell-State (red) as compared to Dynamic Attention LSTM output (green) and the baseline (blue). We see that the model shows faster convergence and stabler training curve indicating a more robust representation of state. In Table \ref{tab:ablation} we compare the performances of each model in Zero-Shot and Multi-Task generalization tasks.

\subsection{Results and Comparison with state-of-the-art}\label{SOTA}

We now show comparison in performance of our Dynamic Attention LSTM - Cell State model with the baseline Gated Attention model in the three difficulty modes and Multi-Task and Zero-Shot generalization settings. We also compare our model with \cite{Misra} which combines the image representation and language representation through concatenation.

 From the curves we observe that the Dynamic Attention model outperforms the baseline in all the difficulty modes in terms of rate of convergence and steady state accuracy. Table \ref{tab_zero_shot} and \ref{tab_multi_task} shows the comparison of Zero-shot and Multi-Task generalization performances with the baseline and concatenation approaches for all three modes of difficulty. We see that our model beats the state of the art by significant margins.This shows that our model has learnt to generalize better in unseen scenarios and with unseen instructions.


\section{Visualization of Attention} \label{attn_viz}

From the attention visualizations Fig. \ref{fig:attentions_viz} we note the following:

\begin{itemize}
    \item At the start of every episode, the attention quickly shifts to the objects leaving the background unattended. It is clear that the agent has learned to detect foreground objects and distinguish it from the background.

\item Even if the field of view is changing constantly, the attention remains fixated on the objects which were under the agent\'s attention. This shows the robustness of the grounding. This is also the case with human attention. We tend to fixate our gaze on the objects that we are observing even though the frame that we are currently seeing is not stationary.
\end{itemize}




\begin{itemize}
    \item The agent quickly manages to focus on the objects of interest based on color description, shape etc as can be seen from the examples. When it fixates on the object(s) of interest, the attention subsides from the other objects. Only the objects very close to the object of interest get some attention as is expected since the agent has to avoid hitting the wrong objects.

\item Fig \ref{fig:attentions_viz}-b depicts a failure case. In this case the attention focuses on the correct object but the agent does not move towards it. This might be to the choice of reward function. Since the agent receives a negative of -0.2 when it approaches an incorrect object and 0 reward when it does not reach any, statistically, not taking any action in certain scenarios might give it a greater expected reward even though correct grounding is achieved.

\end{itemize}


\section{Conclusion}
\label{sec:conclusion}

In this paper we have proposed a dynamic attention network that can ground natural language instructions to visual elements and actions. We showed the effectiveness of the dynamic attention over the static attention model of gated attention network in terms of convergence rate as well as steady state performance. We have shown that the cell-state of an LSTM can be a natural choice for modeling dynamic attention through the performance of A3C and as well as the quality of attention that it generates. We demonstrated the effect of using 1D convolution on the rate of convergence of the network.
Through visualizations we have shown the robustness and quality of the grounding. Finally we conclude that the use of dynamic attention helps in grounding of instructions to objects and actions and is a natural choice when dealing with continuous observation spaces like in a 3D world.
{\small
\bibliographystyle{ieee_fullname}
\bibliography{egbib}

\begin{thebibliography}{10}\itemsep=-1pt

\bibitem{VQA}
Stanislaw Antol, Aishwarya Agrawal, Jiasen Lu, Margaret Mitchell, Dhruv Batra,
  C.~Lawrence Zitnick, and Devi Parikh.
\newblock {VQA}: {V}isual {Q}uestion {A}nswering.
\newblock In {\em International Conference on Computer Vision (ICCV)}, 2015.

\bibitem{Artzi-2013}
Yoav Artzi and Luke Zettlemoyer.
\newblock Weakly supervised learning of semantic parsers for mapping
  instructions to actions.
\newblock {\em Transactions of the Association for Computational Linguistics},
  2013.

\bibitem{Barnard_JMLR2003}
K Barnard, P Duygulu, and D Forsyth.
\newblock N. de freitas, d.
\newblock {\em Blei, and MI Jordan," Matching Words and Pictures", submitted to
  JMLR}, 2003.

\bibitem{Beetz}
M. Beetz, U. Klank, I. Kresse, A. Maldonado, L. Mösenlechner, D. Pangercic, T.
  Rühr, and M. Tenorth.
\newblock Robotic roommates making pancakes.
\newblock In {\em 2011 11th IEEE-RAS International Conference on Humanoid
  Robots}, pages 529--536, Oct 2011.

\bibitem{Bollini2013}
Mario Bollini, Stefanie Tellex, Tyler Thompson, Nicholas Roy, and Daniela Rus.
\newblock {\em Interpreting and Executing Recipes with a Cooking Robot}, pages
  481--495.
\newblock Springer International Publishing, Heidelberg, 2013.

\bibitem{Chao}
C. Chao, M. Cakmak, and A.~L. Thomaz.
\newblock Towards grounding concepts for transfer in goal learning from
  demonstration.
\newblock In {\em 2011 IEEE International Conference on Development and
  Learning (ICDL)}, volume~2, pages 1--6, Aug 2011.

\bibitem{chaplot2017gated}
Devendra~Singh Chaplot, Kanthashree~Mysore Sathyendra, Rama~Kumar Pasumarthi,
  Dheeraj Rajagopal, and Ruslan Salakhutdinov.
\newblock Gated-attention architectures for task-oriented language grounding.
\newblock In {\em Proceedings of the Thirty-Second AAAI Conference on
  Artificial Intelligence(AAAI)}, 2018.

\bibitem{chen:aaai11}
David~L. Chen and Raymond~J. Mooney.
\newblock Learning to interpret natural language navigation instructions from
  observations.
\newblock pages 859--865, August 2011.

\bibitem{Chen_CVPR2015}
Xinlei Chen and C Lawrence~Zitnick.
\newblock Mind's eye: A recurrent visual representation for image caption
  generation.
\newblock In {\em Proceedings of the IEEE conference on computer vision and
  pattern recognition}, pages 2422--2431, 2015.

\bibitem{Das_EMNLP2016}
Abhishek Das, Harsh Agrawal, C.~Lawrence Zitnick, Devi Parikh, and Dhruv Batra.
\newblock {Human Attention in Visual Question Answering: Do Humans and Deep
  Networks Look at the Same Regions?}
\newblock In {\em Conference on Empirical Methods in Natural Language
  Processing (EMNLP)}, 2016.

\bibitem{Das_CVPR2017}
Abhishek Das, Satwik Kottur, Khushi Gupta, Avi Singh, Deshraj Yadav,
  Jos{\'e}~MF Moura, Devi Parikh, and Dhruv Batra.
\newblock Visual dialog.
\newblock In {\em Proceedings of the IEEE Conference on Computer Vision and
  Pattern Recognition}, 2017.

\bibitem{de2017guesswhat}
Harm De~Vries, Florian Strub, Sarath Chandar, Olivier Pietquin, Hugo
  Larochelle, and Aaron Courville.
\newblock Guesswhat?! visual object discovery through multi-modal dialogue.
\newblock In {\em Proc. of CVPR}, 2017.

\bibitem{Fang_CVPR2015}
Hao Fang, Saurabh Gupta, Forrest Iandola, Rupesh Srivastava, Li Deng, Piotr
  Doll{\'a}r, Jianfeng Gao, Xiaodong He, Margaret Mitchell, John Platt, et~al.
\newblock From captions to visual concepts and back.
\newblock In {\em Proceedings of the IEEE conference on computer vision and
  pattern recognition}, 2015.

\bibitem{Farhadi_ECCV2010}
Ali Farhadi, Mohsen Hejrati, Mohammad~Amin Sadeghi, Peter Young, Cyrus
  Rashtchian, Julia Hockenmaier, and David Forsyth.
\newblock Every picture tells a story: Generating sentences from images.
\newblock In {\em European conference on computer vision}, pages 15--29.
  Springer, 2010.

\bibitem{Fukui_arXiv2016}
Akira Fukui, Dong~Huk Park, Daylen Yang, Anna Rohrbach, Trevor Darrell, and
  Marcus Rohrbach.
\newblock Multimodal compact bilinear pooling for visual question answering and
  visual grounding.
\newblock {\em arXiv preprint arXiv:1606.01847}, 2016.

\bibitem{Gao_NIPS2015}
Haoyuan Gao, Junhua Mao, Jie Zhou, Zhiheng Huang, Lei Wang, and Wei Xu.
\newblock Are you talking to a machine? dataset and methods for multilingual
  image question.
\newblock In {\em Advances in Neural Information Processing Systems}, pages
  2296--2304, 2015.

\bibitem{Guadarrama}
S. Guadarrama, L. Riano, D. Golland, D. Go¨hring, Y. Jia, D. Klein, P. Abbeel,
  and T. Darrell.
\newblock Grounding spatial relations for human-robot interaction.
\newblock In {\em 2013 IEEE/RSJ International Conference on Intelligent Robots
  and Systems}, pages 1640--1647, Nov 2013.

\bibitem{Hochreiter}
Sepp Hochreiter and J\"{u}rgen Schmidhuber.
\newblock Long short-term memory.
\newblock {\em Neural Comput.}, 9(8):1735--1780, Nov. 1997.

\bibitem{jain_CVPR2017}
Unnat Jain, Ziyu Zhang, and Alexander Schwing.
\newblock Creativity: Generating diverse questions using variational
  autoencoders.
\newblock {\em arXiv preprint arXiv:1704.03493}, 2017.

\bibitem{Johnson_CVPR2016}
Justin Johnson, Andrej Karpathy, and Li Fei-Fei.
\newblock Densecap: Fully convolutional localization networks for dense
  captioning.
\newblock In {\em Proceedings of the IEEE Conference on Computer Vision and
  Pattern Recognition}, pages 4565--4574, 2016.

\bibitem{Karpathy_CVPR2015}
Andrej Karpathy and Li Fei-Fei.
\newblock Deep visual-semantic alignments for generating image descriptions.
\newblock In {\em Proceedings of the IEEE conference on computer vision and
  pattern recognition}, pages 3128--3137, 2015.

\bibitem{KempkaWRTJ16}
Michal Kempka, Marek Wydmuch, Grzegorz Runc, Jakub Toczek, and Wojciech
  Jaskowski.
\newblock Vizdoom: {A} doom-based {AI} research platform for visual
  reinforcement learning.
\newblock {\em CoRR}, abs/1605.02097, 2016.

\bibitem{Kulkarni_CVPR2011}
Girish Kulkarni, Visruth Premraj, Sagnik Dhar, Siming Li, Yejin Choi,
  Alexander~C Berg, and Tamara~L Berg.
\newblock Baby talk: Understanding and generating image descriptions.
\newblock In {\em Proceedings of the 24th CVPR}. Citeseer, 2011.

\bibitem{LeCunn}
Yann Lecun and Yoshua Bengio.
\newblock {\em Convolutional networks for images, speech, and time-series}.
\newblock MIT Press, 1995.

\bibitem{Lemaignan2012}
S{\'e}verin Lemaignan, Raquel Ros, E.~Akin Sisbot, Rachid Alami, and Michael
  Beetz.
\newblock Grounding the interaction: Anchoring situated discourse in everyday
  human-robot interaction.
\newblock {\em International Journal of Social Robotics}, 4(2):181--199, Apr
  2012.

\bibitem{Lin_ECCV2014}
Tsung-Yi Lin, Michael Maire, Serge Belongie, James Hays, Pietro Perona, Deva
  Ramanan, Piotr Doll{\'a}r, and C~Lawrence Zitnick.
\newblock Microsoft coco: Common objects in context.
\newblock In {\em European Conference on Computer Vision}, pages 740--755.
  Springer, 2014.

\bibitem{Lu_NIPS2016}
Jiasen Lu, Jianwei Yang, Dhruv Batra, and Devi Parikh.
\newblock Hierarchical question-image co-attention for visual question
  answering.
\newblock In {\em Advances In Neural Information Processing Systems}, pages
  289--297, 2016.

\bibitem{Ma_AAAI2016}
Lin Ma, Zhengdong Lu, and Hang Li.
\newblock Learning to answer questions from image using convolutional neural
  network.
\newblock In {\em Thirtieth AAAI Conference on Artificial Intelligence}, 2016.

\bibitem{Malinowski_NIPS2014}
Mateusz Malinowski and Mario Fritz.
\newblock A multi-world approach to question answering about real-world scenes
  based on uncertain input.
\newblock In {\em Advances in Neural Information Processing Systems (NIPS)},
  2014.

\bibitem{Mei}
Hongyuan Mei, Mohit Bansal, and Matthew~R. Walter.
\newblock Listen, attend, and walk: Neural mapping of navigational instructions
  to action sequences.
\newblock {\em CoRR}, abs/1506.04089, 2015.

\bibitem{Misra}
Dipendra~Kumar Misra, John Langford, and Yoav Artzi.
\newblock Mapping instructions and visual observations to actions with
  reinforcement learning.
\newblock {\em CoRR}, abs/1704.08795, 2017.

\bibitem{pmlr-v48-mniha16}
Volodymyr Mnih, Adria~Puigdomenech Badia, Mehdi Mirza, Alex Graves, Timothy
  Lillicrap, Tim Harley, David Silver, and Koray Kavukcuoglu.
\newblock Asynchronous methods for deep reinforcement learning.
\newblock In {\em Proceedings of The 33rd International Conference on Machine
  Learning}, pages 1928--1937, 2016.

\bibitem{Mostafazadeh_ACL2016}
Nasrin Mostafazadeh, Ishan Misra, Jacob Devlin, Margaret Mitchell, Xiaodong He,
  and Lucy Vanderwende.
\newblock Generating natural questions about an image.
\newblock {\em arXiv preprint arXiv:1603.06059}, 2016.

\bibitem{Noh_CVPR2016}
Hyeonwoo Noh, Paul Hongsuck~Seo, and Bohyung Han.
\newblock Image question answering using convolutional neural network with
  dynamic parameter prediction.
\newblock In {\em Proceedings of the IEEE Conference on Computer Vision and
  Pattern Recognition}, pages 30--38, 2016.

\bibitem{OhSLK17}
Junhyuk Oh, Satinder~P. Singh, Honglak Lee, and Pushmeet Kohli.
\newblock Zero-shot task generalization with multi-task deep reinforcement
  learning.
\newblock {\em CoRR}, abs/1706.05064, 2017.

\bibitem{Patro_CVPR2018dvqa}
Badri Patro and Vinay~P. Namboodiri.
\newblock Differential attention for visual question answering.
\newblock In {\em The IEEE Conference on Computer Vision and Pattern
  Recognition (CVPR)}, June 2018.

\bibitem{Patro2019ProbabilisticFF}
Badri~N. Patro, Anupriy, and Vinay~P. Namboodiri.
\newblock Probabilistic framework for solving visual dialog.
\newblock {\em ArXiv}, abs/1909.04800, 2019.

\bibitem{Patro_EMNLP2018MDN}
Badri~Narayana Patro, Sandeep Kumar, Vinod~Kumar Kurmi, and Vinay Namboodiri.
\newblock Multimodal differential network for visual question generation.
\newblock In {\em Proceedings of the 2018 Conference on Empirical Methods in
  Natural Language Processing}, pages 4002--4012. Association for Computational
  Linguistics, 2018.

\bibitem{Patro_COLING2018learning}
Badri~Narayana Patro, Vinod~Kumar Kurmi, Sandeep Kumar, and Vinay Namboodiri.
\newblock Learning semantic sentence embeddings using sequential pair-wise
  discriminator.
\newblock In {\em Proceedings of the 27th International Conference on
  Computational Linguistics}, pages 2715--2729, 2018.

\bibitem{Patro_ICCV2019}
Badri~N. Patro, Mayank Lunayach, Shivansh Patel, and Vinay~P. Namboodiri.
\newblock U-cam: Visual explanation using uncertainty based class activation
  maps.
\newblock In {\em arXiv preprint arXiv:1908.06306}, 2019.

\bibitem{Ren_NIPS2015}
Mengye Ren, Ryan Kiros, and Richard Zemel.
\newblock Exploring models and data for image question answering.
\newblock In {\em Advances in Neural Information Processing Systems (NIPS)},
  pages 2953--2961, 2015.

\bibitem{Shih_CVPR2016}
Kevin~J Shih, Saurabh Singh, and Derek Hoiem.
\newblock Where to look: Focus regions for visual question answering.
\newblock In {\em Proceedings of the IEEE Conference on Computer Vision and
  Pattern Recognition}, pages 4613--4621, 2016.

\bibitem{Socher_TACL2014}
Richard Socher, Andrej Karpathy, Quoc~V Le, Christopher~D Manning, and Andrew~Y
  Ng.
\newblock Grounded compositional semantics for finding and describing images
  with sentences.
\newblock {\em Transactions of the Association of Computational Linguistics},
  2(1):207--218, 2014.

\bibitem{strub2017end}
Florian Strub, Harm De~Vries, Jeremie Mary, Bilal Piot, Aaron Courville, and
  Olivier Pietquin.
\newblock End-to-end optimization of goal-driven and visually grounded dialogue
  systems.
\newblock {\em arXiv preprint arXiv:1703.05423}, 2017.

\bibitem{Vinyals_CVPR2015}
Oriol Vinyals, Alexander Toshev, Samy Bengio, and Dumitru Erhan.
\newblock Show and tell: A neural image caption generator.
\newblock In {\em Proceedings of the IEEE Conference on Computer Vision and
  Pattern Recognition}, pages 3156--3164, 2015.

\bibitem{Xu_ECCV2016}
Huijuan Xu and Kate Saenko.
\newblock Ask, attend and answer: Exploring question-guided spatial attention
  for visual question answering.
\newblock In {\em European Conference on Computer Vision}, pages 451--466.
  Springer, 2016.

\bibitem{Xu_ICML2015}
Kelvin Xu, Jimmy Ba, Ryan Kiros, Kyunghyun Cho, Aaron Courville, Ruslan
  Salakhudinov, Rich Zemel, and Yoshua Bengio.
\newblock Show, attend and tell: Neural image caption generation with visual
  attention.
\newblock In {\em International Conference on Machine Learning}, pages
  2048--2057, 2015.

\bibitem{Yan_ECCV2016}
Xinchen Yan, Jimei Yang, Kihyuk Sohn, and Honglak Lee.
\newblock Attribute2image: Conditional image generation from visual attributes.
\newblock In {\em European Conference on Computer Vision}, pages 776--791.
  Springer, 2016.

\bibitem{DBLP:YuZX17}
Haonan Yu, Haichao Zhang, and Wei Xu.
\newblock A deep compositional framework for human-like language acquisition in
  virtual environment.
\newblock {\em CoRR}, abs/1703.09831, 2017.

\bibitem{Zhu_CVPR2016}
Yuke Zhu, Oliver Groth, Michael Bernstein, and Li Fei-Fei.
\newblock Visual7w: Grounded question answering in images.
\newblock In {\em Proceedings of the IEEE Conference on Computer Vision and
  Pattern Recognition}, pages 4995--5004, 2016.

\end{thebibliography}
}

\end{document}